\documentclass[conference]{IEEEtran}
\IEEEoverridecommandlockouts
\usepackage{cite}
\usepackage{amsmath,amssymb,amsfonts}
\usepackage{fancyhdr}
\usepackage{algorithmic}
\usepackage{graphicx}
\usepackage{textcomp}
\usepackage{xcolor}
\usepackage{float}
\usepackage{dblfloatfix}
\def\BibTeX{{\rm B\kern-.05em{\sc i\kern-.025em b}\kern-.08em
    T\kern-.1667em\lower.7ex\hbox{E}\kern-.125emX}}

\makeatletter
\let\old@ps@headings\ps@headings
\let\old@ps@IEEEtitlepagestyle\ps@IEEEtitlepagestyle
\def\confheader#1{%
  \def\ps@headings{%
    \old@ps@headings%
    \def\@oddhead{\strut\hfill#1\hfill\strut}%
    \def\@evenhead{\strut\hfill#1\hfill\strut}%
  }%
  \def\ps@IEEEtitlepagestyle{%
    \old@ps@IEEEtitlepagestyle%
    \def\@oddhead{\strut\hfill#1\hfill\strut}%
    \def\@evenhead{\strut\hfill#1\hfill\strut}%
  }%
  \ps@headings%
}
\makeatother

\begin{document}

\title{NerfBridge: Bringing Real-time, Online Neural Radiance Field Training to Robotics\\
\thanks{*Corresponding Author: \texttt{javieryu@stanford.edu}. \\ 
1. Stanford University Department of Aeronautics and Astronautics \\
2. Stanford University Department of Mechanical Engineering. \\
This work was funded in part by ONR grant N00014-18-1-2830, DARPA grant HR001120C0107, and a gift from Applied Intuition, Inc.}
}
\confheader{Presented at the ICRA 2023 Workshop on Unconventional Spatial Representations}
\author{\IEEEauthorblockN{Javier Yu$^{1*}$}
\and
\IEEEauthorblockN{Jun En Low$^2$}
\and 
\IEEEauthorblockN{Keiko Nagami$^1$}
\and
\IEEEauthorblockN{Mac Schwager$^1$}
}

\maketitle

\begin{abstract}
Neural radiance fields (NeRFs) are a class of implicit scene representations that model 3D environments from color images. NeRFs are expressive, and can model the complex and multi-scale geometry of real world environments, which potentially makes them a powerful tool for robotics applications. Modern NeRF training libraries can generate a photo-realistic NeRF from a static data set in just a few seconds, but are designed for offline use and require a slow pose optimization pre-computation step.

In this work we propose NerfBridge, an open-source bridge between the Robot Operating System (ROS) and the popular Nerfstudio library for real-time, online training of NeRFs from a stream of images. NerfBridge enables rapid development of research on applications of NeRFs in robotics by providing an extensible interface to the efficient training pipelines and model libraries provided by Nerfstudio. As an example use case we outline a hardware setup that can be used NerfBridge to train a NeRF from images captured by a camera mounted to a quadrotor in both indoor and outdoor environments.

For accompanying video https://youtu.be/EH0SLn-RcDg and code https://github.com/javieryu/nerf\_bridge.
\end{abstract}


\section{Introduction}
Neural implicit scene representations offer an expressive and memory efficient alternative to traditional discrete scene representations like voxels or point clouds. One class of these implicit representations are Neural Radiance Fields (NeRF) which, in their most basic form, use a data set of color image and camera pose pairs to supervise the training of a neural network which in turn learns a continuous map of the environment captured in the data set's images. The relative simplicity and flexibility of NeRF-based representations has the potential to change the way that 3D environments are represented for robotics applications.

Nerfstudio \cite{nerfstudio} is a modular library for NeRF development, and provides easy access to efficient implementations of state-of-the-art NeRF training pipelines and models. However, it is in large part designed for offline applications where data is gathered in entirety prior to training the NeRF. For applications in robotics this workflow is not easily adaptable because for those problems data is continuously received as a stream from the robot's various onboard sensors. Typically, these onboard sensors and downstream tasks are orchestrated using the Robot Operating System (ROS) \cite{quigley2009ros}. To that end, to make integration and development as seamless as possible, we propose NerfBridge, a software package that bridges the gap between NerfStudio and ROS.

No two robotics platforms have the same requirements, and so our goal with NerfBridge is not to provide a package that is one size fits all. Instead we developed a minimal and adaptable interface between the two libraries that practitioners can use as a foundation for their application specific uses. 

Work related to online NeRF training is covered in Section \ref{related}, the basic functionality of NerfBridge is outlined in Section \ref{outline}, and then in Section \ref{hardware} we provide a detailed description of how a camera equipped quadrotor and a ground station can be used to construct a NeRF in real-time. Finally, in Section \ref{conclusion} we discuss potential research directions at the intersection of robotics and neural implicit scene representations.

\begin{figure}
    \centering
    \includegraphics[width=\linewidth]{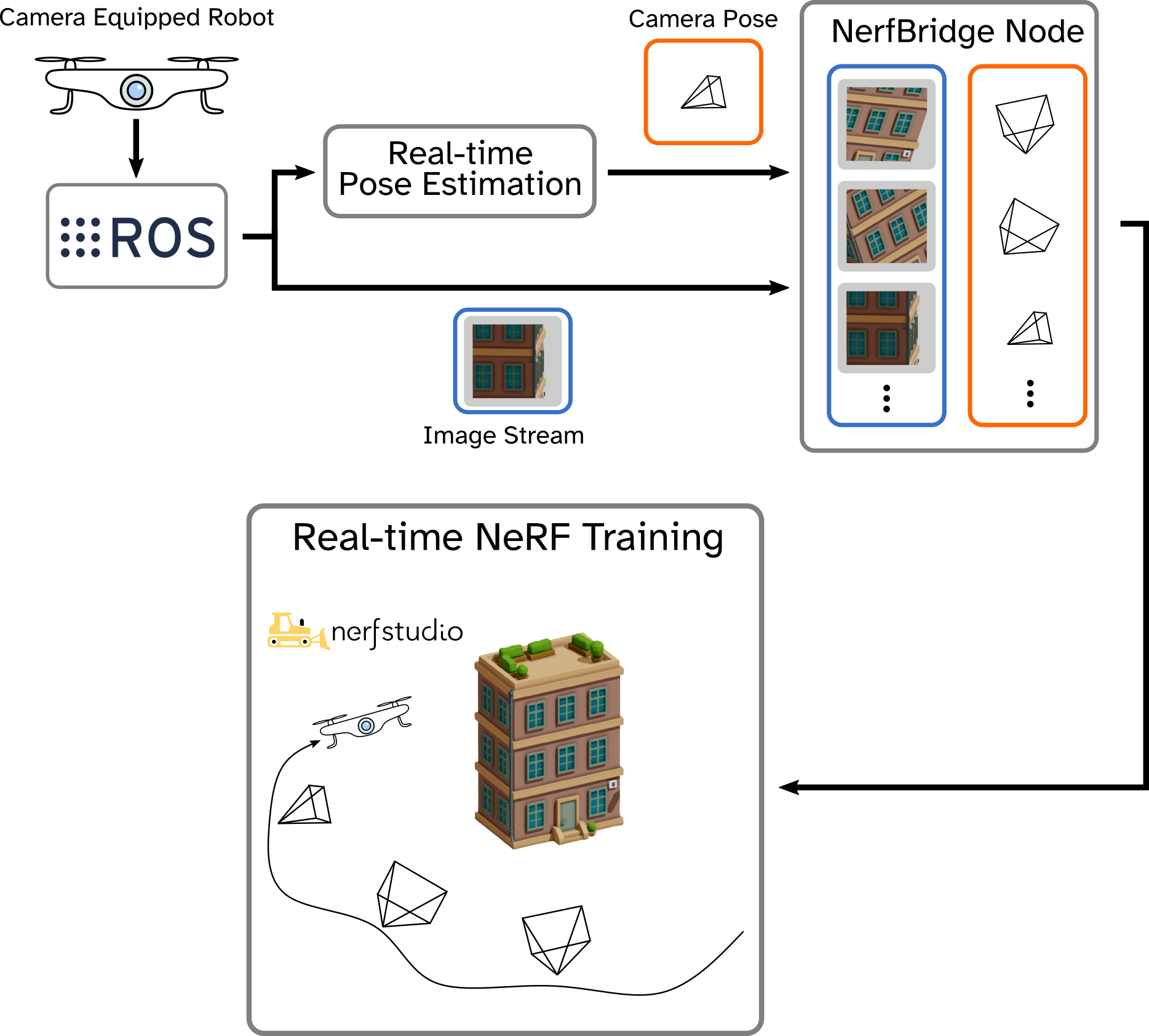}
    \caption{A basic outline of the functionality of the NerfBridge Package for integrating streaming images with real-time NeRF training.}
    \label{fig:schem}
\end{figure}

\section{Related Work}
\label{related}
Early work with NeRF required training times of at least an hour, but often longer, to achieve a NeRF with sufficient quality to be used in down-stream robotics tasks \cite{mildenhall2021nerf}. However, the ground-breaking work in \cite{muller2022instant} demonstrated that, using a number of innovations and optimizations, NeRF training times could be reduced to just a few seconds.

The potential for NeRF as a spatial representation in a Simultaneous Localization and Mapping (SLAM) algorithm was first demonstrated in \cite{sucar2021imap} where a neural implicit map and camera poses are jointly optimized using RGB-D images. Later work, \cite{zhu2022nice}, showed that using hierarchical NeRF architectures improved the reconstruction accuracy of the environment. Unlike the two previous works that use RGB-D images as input, \cite{rosinol2022nerf} uses RGB images, and the outputs of a dense SLAM algorithm to build a NeRF map, and demonstrates that this enables higher fidelity implicit maps. 

All of the methods \cite{sucar2021imap}, \cite{zhu2022nice}, and \cite{rosinol2022nerf} offer variations on a similar idea, but none of them are particularly well suited for integration with existing robotics platforms because they lack existing open-source implementations with ROS. Furthermore, the existing code for these implementations lacks modularity, and restricts the user to the NeRF architectures and pose estimation methods selected by the authors. With NerfBridge, users are free to choose their NeRF architecture from the numerous methods already implemented in Nerfstudio, and can use any pose estimation method that is compatible with ROS.

\section{NerfBridge}
\label{outline}
Traditional NeRF training requires two inputs a color image and the pose of the camera used to take that image. Using the intrinsic parameters of the camera, a NeRF is generated by supervising it's underlying neural network using a ray-tracing reconstruction loss \cite{mildenhall2021nerf}.

For online training of a NeRF it is therefore necessary to provide access to a stream of posed color images, and at initialization intrinsic parameters for the camera that is being used. Since NerfBridge is designed to work with ROS these values are passed as messages that are published to independent ROS topics --- one topic for pose and one topic for images. At it's core NerfBridge creates a ROS node that listens to these topics, and continuously inserts new images and poses into pre-allocated arrays. In parrallel, Nerfstudio is used to continuously train and update a NeRF using pixels from the available pool of images. This process continues until the training arrays have been filled at which point no new images are added, and NeRF training proceeds on the static data set until convergence.

The task of estimating the poses of each image is often overlooked in the NeRF literature, and offline NeRF approaches typically use the structure from motion package COLMAP \cite{schoenberger2016sfm} to assign poses to entire image data sets. In a streaming context this is not a viable option, and instead poses must be computed in real-time. The poses required for NeRF training can be estimated in a number of different ways including external motion tracking systems, and visual odometry methods. In our hardware implementation we use the open-source, visual odometry package ORBSLAM3 \cite{ORBSLAM3_TRO} to estimate the camera poses. 

Part of the design philosophy of NerfBridge is to limit it to essential functionality rather than attempting to make it feature rich, and thus making it easier to maintain and faster to adapt to new applications. To that end, we do not implement possible extensions (ex. key-framing) to maintain the simplicity of NerfBridge, and because, in large part, online NeRF training is a relatively unexplored field and the benefit of these extensions have yet to be studied.

\section{Mapping Case Study}
\label{hardware}
One basic application of NeRF is mapping, and in this section we outline how a camera equipped quadrotor and a computing ground station can be coordinated using NerfBridge to build a NeRF of an object of interest. Training the NeRF in real-time allows the operator to use the current quality of the NeRF as feedback while the quadrotor is being flown through the region of interest. This avoids the offline training workflow where the operator would have to land the quadrotor, offload the captured images, train the NeRF, and then re-deploy for more images if the quality of the NeRF is poor.

In this implementation, the quadrotor sends images over a WiFi connection to the ground station computer. The ground station then uses ORBSLAM3 \cite{ORBSLAM3_TRO} to estimate the pose of the camera at each frame, and this pose and image pair are in turn processed and passed to Nerfstudio via NerfBridge.

\subsection{Hardware Details}
The quadrotor's on-board computer is a Raspberry Pi 4B running Ubuntu 20.04 and ROS Noetic, and is used to operate the camera and communicate with the ground station via WiFi.

Arguably the most important piece of hardware for this application is the camera, and in this case we chose a 1.2 MP global shutter USB camera (oCam-1GNN-U) from WithRobot. The main consideration here being that global shutter cameras have fewer image artifacts like motion blur, and additionally the producer of this camera provides a publicly available ROS Node implementations that means integration with an existing robotics platform is relatively straight-forward.

A ground station with an Nvidia GPU is also essential for real-time training because Nerfstudio uses CUDA \cite{kirk2007nvidia} to optimize the NeRF training pipeline. In our setup, we use a desktop computer with an RTX 3090 GPU, AMD Ryzen 9 5900X CPU, and 32 GB of RAM which provides more than enough compute to run ORBSLAM3, NerfBridge, and monitoring software in parallel. 

\subsection{Indoor Flight Details}
The first test of our setup was a mock indoor mapping scenario in which the quadrotor flew a helical trajectory around a foam pillar and set of pipes. We use motion capture cameras to provide position information for our flight controller, and separately use visual odometry to estimate the poses for NerfBridge. During the mission, the quadrotor streams images at approximately 20 Hz, and these are sub-sampled at 2 Hz by NerfBridge. The flight time was roughly 2.5 minutes, and resulted in a final image set of about 300 images.

\begin{figure}[h]
    \centering
    \includegraphics[width=0.8\linewidth]{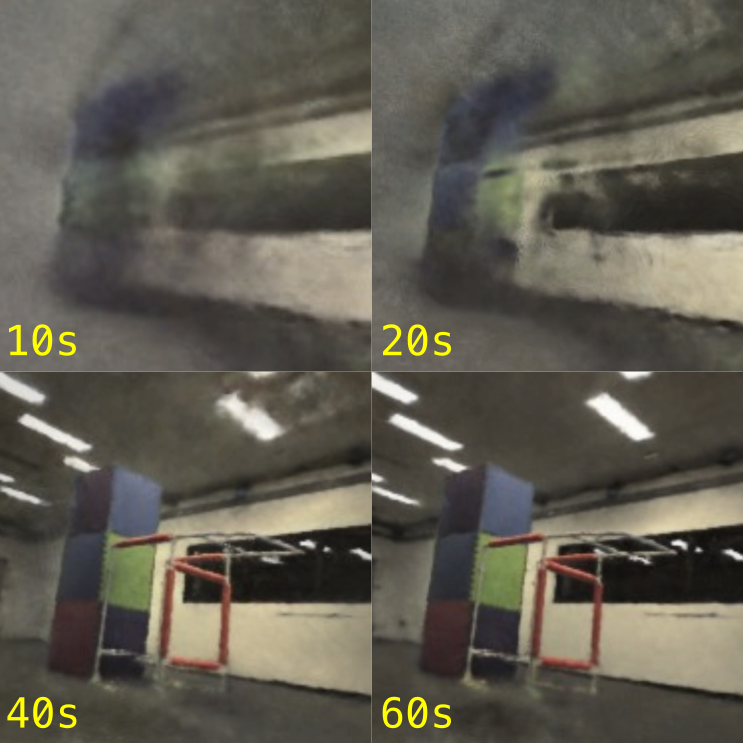}
    \caption{The reconstruction of the NeRF generated overtime from flying a helical trajectory around a foam box and pipes using NerfBridge.}
    \label{fig:indoor}
\end{figure}

Figure \ref{fig:indoor} shows a progression of the NeRF quality as more images are added, and the quadrotor is flying. After about a minute of flight time the newly added images are largely redundant, and do not result in substantial improvements in NeRF map quality. The final NeRF includes both an accurate reconstruction of the object of interest (pillar), but also the surrounding room including windows, lights, and glass.

\subsection{Outdoor Flight Details}
To verify that our setup can also work in more realistic, outdoor conditions we also tested on a outdoor mapping scenario in which the quadrotor flew a raster trajectory at close to ground level with the point of interest being the side of a building. This flight was conducted at the Elliot Center on Stanford University Campus. In this case, a GPS and onboard sensors are used to maintain stable flight, and visual odometry is again separately used to estimate poses for NerfBridge. Flight times and sampling rates are the same from the indoor experiment.

\begin{figure}[h]
    \centering
    \includegraphics[width=\linewidth]{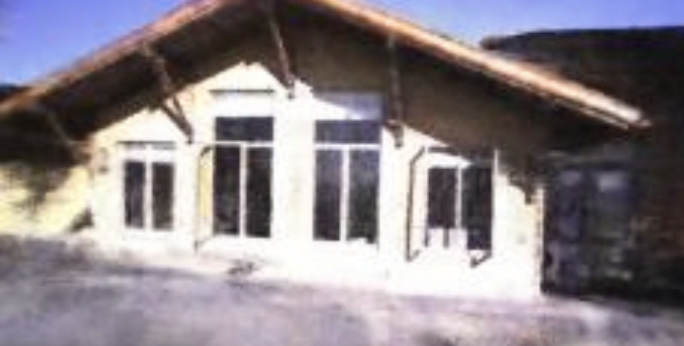}
    \caption{The reconstruction of the NeRF generated using NeRF Bridge for outdoor mapping at the Elliot Center on Stanford University Campus.}
    \label{fig:outdoor}
\end{figure}

In Figure \ref{fig:outdoor} is a rendering of the resulting NeRF. NerfBridge is able to capture the multi-scale structures of the building facade and windows.

\section{Conclusions and Future Work}
\label{conclusion}
The core objective of NerfBridge is to streamline the process for integrating neural implicit maps in robotics pipelines, and accelerate exploration of applications of NeRFs in robotics. To that end we designed a modular, ROS-based software package that can interface state-of-the-art NeRF training libraries with existing robotics platforms.

In future work, we hope to combine NeRF navigation algorithms \cite{adamkiewicz2022vision} with online NeRF training as a novel modality for robot trajectory optimization and mapping. Online NeRF training itself is also a relatively unexplored field, and NerfBridge opens the opportunity to study the effects of novel, information-based keyframing schemes to avoid catastrophic forgetting during NeRF training.

\bibliographystyle{IEEEtran}
\bibliography{refs.bib}

\end{document}